\pdfoutput=1

\documentclass[11pt]{article}

\usepackage{EMNLP2023}

\usepackage{times}
\usepackage{latexsym}

\usepackage[T1]{fontenc}

\usepackage[utf8]{inputenc}

\usepackage{microtype}

\usepackage{amssymb}
\usepackage{amsmath}
\usepackage{graphicx}
\usepackage{subfigure}

\usepackage{booktabs}
\usepackage{multirow}

\usepackage{inconsolata}

\title{\vspace{-1em}\emph{BeautifulPrompt}: Towards Automatic Prompt Engineering for Text-to-Image Synthesis}

\author{
    Tingfeng Cao$^{1,2,3}$\thanks{\ \ Work done during an internship at Alibaba.},\ 
    Chengyu Wang$^{2}$\thanks{\ \ C. Wang and J. Zhu are co-corresponding authors.},\ 
    Bingyan Liu$^{1,2}$,\ 
    Ziheng Wu$^{2}$,\\
    \textbf{Jinhui Zhu}$^{1,3\dagger}$,\ 
    \textbf{Jun Huang}$^{2}$\\
    $^1$South China University of Technology, China\\
    $^2$Alibaba Group, China\\
    $^3$Key Laboratory of Big Data and Intelligent Robot (South China University of Technology) \\
    Ministry of Education, China\\
    \normalsize
    \texttt{\{setingfengcao, eeliubingyan\}@mail.scut.edu.cn, csjhzhu@scut.edu.cn} \\
    \normalsize
    \texttt{\{chengyu.wcy, zhoulou.wzh, huangjun.hj\}@alibaba-inc.com}
}

\begin{document}
\maketitle
\begin{abstract}
Recently, diffusion-based deep generative models (e.g., Stable Diffusion) have shown impressive results in text-to-image synthesis. However, current text-to-image models often require multiple passes of prompt engineering by humans in order to produce satisfactory results for real-world applications.
We propose~\emph{BeautifulPrompt}, a deep generative model to produce high-quality prompts from very simple raw descriptions, which
enables diffusion-based models to generate more beautiful images.
In our work, we first fine-tuned the~\emph{BeautifulPrompt} model over low-quality and high-quality collecting prompt pairs.
Then, to ensure that our generated prompts can generate more beautiful images, we further propose a Reinforcement Learning with Visual AI Feedback technique to fine-tune our model to maximize the reward values of the generated prompts, where the reward values are calculated based on the PickScore and the Aesthetic Scores.
Our results demonstrate that learning from visual AI feedback promises the potential to improve the quality of generated prompts and images significantly. We further showcase the integration of~\emph{BeautifulPrompt} to a cloud-native AI platform to provide better text-to-image generation service in the cloud.
\footnote{Datasets and source codes will be publicly available in the EasyNLP framework~\cite{easynlp}. URL: \url{https://github.com/alibaba/EasyNLP}.
Models are released in HuggingFace under the names: pai-bloom-1b1-text2prompt-sd (\url{https://huggingface.co/alibaba-pai/pai-bloom-1b1-text2prompt-sd}) and pai-bloom-1b1-text2prompt-sd-v2 (\url{https://huggingface.co/alibaba-pai/pai-bloom-1b1-text2prompt-sd-v2}), where pai-bloom-1b1-text2prompt-sd is the model introduced in this work, and pai-bloom-1b1-text2prompt-sd-v2 is the enhanced version trained with a lareger dateset.}

\end{abstract}

\section{Introduction}

\begin{figure*}[ht]
\centering
\includegraphics[width=0.95\textwidth, keepaspectratio]
{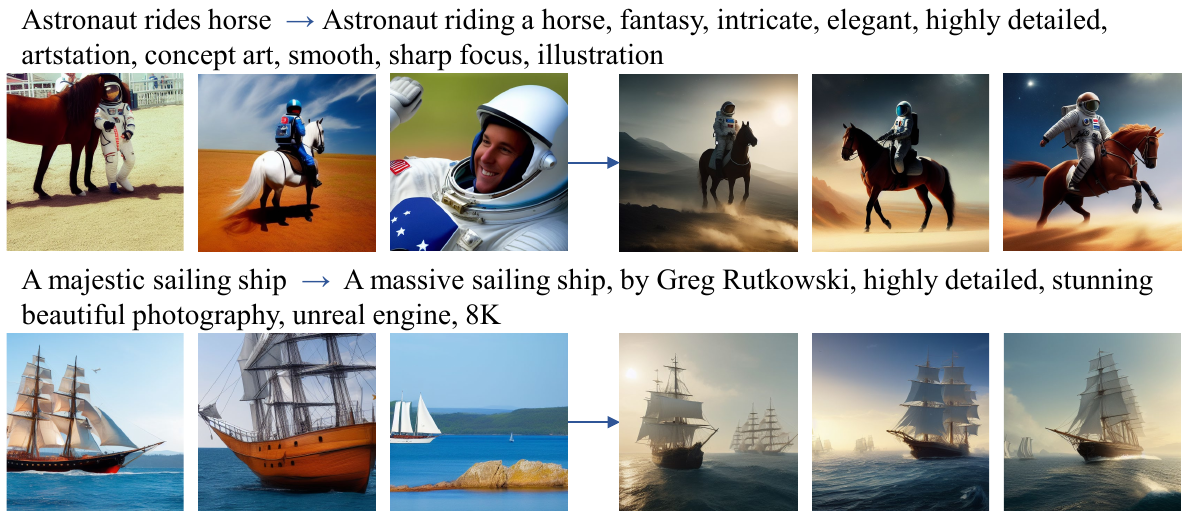}
\caption{Comparing the qualities of images generated from the original prompts (left) with those from the prompts generated by~\emph{BeautifulPrompt} (right). The underlying TIS model is Stable Diffusion 1.5.}
\label{imgs}
\end{figure*}

Text-to-Image Synthesis (TIS) is one of the most spectacularly developed and widely applied techniques in generative Artificial Intelligence (AI), aiming to create realistic images with texts as input. Recently, with the advance of the modeling power of large models, TIS is undergoing a revolution. Large-scale TIS models, such as DALLE~\cite{ramesh2021zero},  DALLE-2~\cite{ramesh2022hierarchical}, latent diffusion models~\cite{rombach2022high} and Imagen~\cite{saharia2022photorealistic}, significantly improve the state-of-the-art performance and allow users without artistic expertise to create unprecedented images through personal imagination.

Yet, TIS models require users to write text prompts before model inference (e.g.,~``A majestic sailing ship'').
Writing such prompts that meet the designer's or art worker's needs is full of uncertainty, like opening a surprise box~\cite{oppenlaender2022taxonomy, design}. 
This is due to the quality of the training data, leading to the need for detailed descriptions to produce high-quality images.
In real-world scenarios, non-experts often find it difficult to write these prompts, and
need to do iterative modification through trials and errors to re-generate the images, leading to a significant loss of time and computing resources.

Prompt engineering is an emerging research field, aiming to explore how to provide prompts for deep generative models and improve the efficiency of direct interaction between humans and AI~\cite{oppenlaender2022taxonomy}.
For example, a user can give a task-oriented prompt and ask ChatGPT~\cite{chatgpt} to generate texts according to the prompt. 
For TIS, the user can write a simple prompt and then ask ChatGPT to supplement the contents. However, directly using ChatGPT to write prompts falls into the dilemma of generating irrelevant and plausible images. 
Hence, the generated prompts can be better in quality if the underlying language model is optimized for the task.
We can see that fine-tuning a language model such as~\cite{brown2020language, bloom, llama} for TIS prompt generation will be a more worthwhile exploration.


In this paper, we propose a new generative model that can write high-quality prompts for diffusion-based models, named \emph{BeautifulPrompt}.
For better user experience,
it re-writes and optimizes the original, low-quality prompts into high-qualities ones to generate better images. It also provides a good source of inspiration for further manual prompt editing. 
Specifically, we first collect a dataset for training \emph{BeautifulPrompt} using an automated data collection pipeline based on existing AI models.
The dataset is used for supervised fine-tuning.
We further propose a Reinforcement Learning with Visual AI Feedback (RLVAIF) technique to maximize the reward values of the generated prompts, which are determined by a couple of trained reward models based on visual signals. The gradient update process of RLVAIF makes the generated prompts more compatible with human preferences without any manual labeling.
A simple comparison of prompts and the resulting images are shown in Figure~\ref{imgs}.
In summary, the main contributions of this study are as follows:

\begin{itemize}

\item We release a new dataset containing 143k prompt pairs and 2k test prompts, enabling researchers to develop prompt engineering models for their TIS applications.

\item We propose~\emph{BeautifulPrompt},  a novel generative model that can write high-quality prompts for diffusion-based TIS models.
A Reinforcement Learning with Visual AI Feedback training scheme is further proposed for better visual alignment without human labeling.


\item Extensive experimental results show the superiority of~\emph{BeautifulPrompt} over strong baselines. We further showcase the integration of~\emph{BeautifulPrompt} to an industrial product to provide better image generation service.
\end{itemize}

\section{Related Work}

\subsection{Text-to-Image Synthesis (TIS)}
TIS is a multi-modal task of generating images conditioned on texts. In the early years, popular image generation networks were mainly based on Generative Adversarial Network (GAN)~\cite{GAN,reed2016generative}. 
Recently, diffusion models~\cite{ho2020denoising, sohl2015deep,DBLP:conf/acl/LiuLD0WZJJC023}, such as DALLE-2~\cite{ramesh2022hierarchical}, Imagen~\cite{saharia2022photorealistic}, and Stable Diffusion~\cite{rombach2022high} have achieved remarkable results. 
Yet, the qualities of generated images depend on prompts. 
In this paper, we propose a prompt generation model, dedicated to optimizing input prompts to generate more beautiful images.

\subsection{TIS Evaluation}
There are several metrics for evaluating TIS. CLIP score~\cite{clip} measures the similarity between generated images and prompts. Aesthetic score~\cite{aesthetic} evaluates the aesthetic quality  of individual images. There are also metrics trained to align with human preferences, such as HPS~\cite{hps}, Image Reward~\cite{imagereward}, and PickScore~\cite{pickscore}. Human preferences can be complex and may involve various dimensions, including the similarity between text and images, as well as image fidelity, aesthetics, and other factors. These evaluation metrics can all serve as visual feedback to optimize the training of prompt engineering models. Among the human preference metrics, PickScore stands out due to its stable scoring and larger, more diverse training datasets, which includes a wider range of implementations (e.g., model size, backbone, hyperparameters)~\cite{pickscore}. These factors can potentially contribute to more stable training and facilitate easier extension to other TIS models.

\subsection{Prompt Engineering for TIS}
Due to the extraordinary potential of TIS, there is a surge of interest in prompt engineering (i.e., creating good prompts). \citet{design} conduct a series of experiments and propose several design guidelines for text-to-image prompt engineering. \citet{oppenlaender2022taxonomy} identifies six types of prompt modifiers through a three-month ethnographic study of the online generative art community. However, these studies are limited to the long and tedious manual prompt engineering. 

BestPrompt~\cite{bestprompt} uses a genetic algorithm to detect keywords to form prompts in order to achieve the best images aesthetically. MagicPrompt\footnote{\url{https://huggingface.co/Gustavosta/MagicPrompt-Stable-Diffusion}} is a popular automatic prompt completion model trained from good prompts collected from the Internet. But these models only serve to complete the prompts.
\emph{BeautifulPrompt}, on the other hand, can re-write the original prompts to give users a good source of inspiration and generate more beautiful images.


\section{Dataset Creation}


In this section, we show the detailed data collection process for~\emph{BeautifulPrompt} training.

\begin{figure}[ht]
\centering
\includegraphics[width=.5\textwidth, keepaspectratio]
{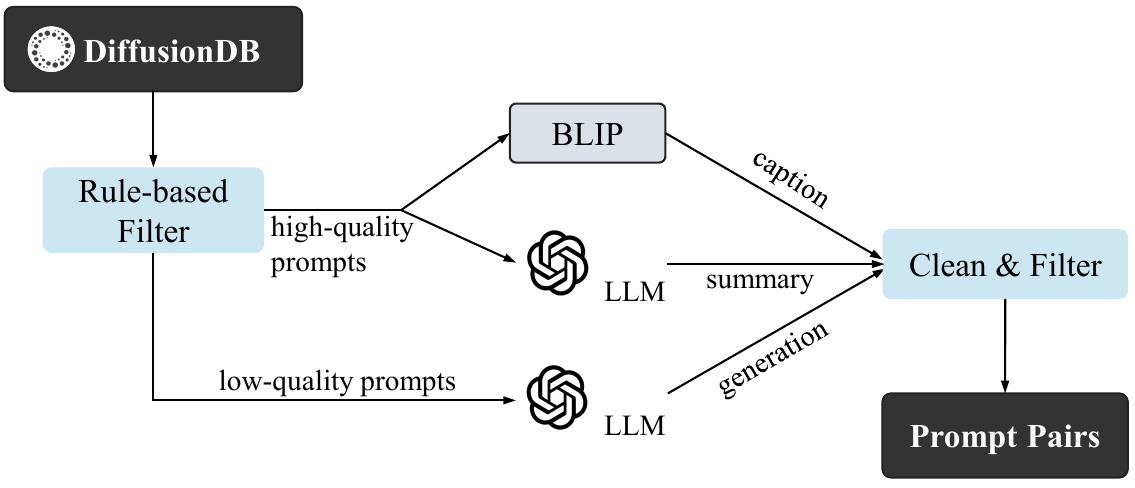}
\caption{The data collection process.}
\label{data}
\end{figure}

\noindent\textbf{Collection of Prompt Pairs.}
The goal of this step is collecting pairs of high-quality and low-quality prompts with similar semantics.
As shown in Figure~\ref{data}, the original data source is DiffusionDB~\cite{diffusiondb}, which contains un-paired prompts only. Heuristically, we split the prompts into low-quality and high-quality ones according to the length of the prompts, the certain tags contained in the prompts, etc. 
Next, we 
i) use BLIP~\cite{blip} to caption the images associated with high-quality prompts and treat the results as the corresponding low-quality prompts, as the captions are shorter and lack details;  
ii) use ChatGPT to summarize the high-quality prompts and treat the summaries as low-quality prompts;
iii) use ChatGPT to generate better prompts from low-quality prompts; the results are considered high-quality prompts.\footnote{The prompts and examples for invoking ChatGPT can be found in Appendix~\ref{chatgpt}.}
Through the above three approaches, we obtain a large number of prompt pairs; however, the quality of these prompt pairs cannot be guaranteed. Hence, we need to do further data cleaning and filtering.

\noindent\textbf{Post-processing.}
We first filter out the examples that are non-English and NSFW (Not Safe For Work). Next, we filter out examples of images generated from high-quality prompts with low aesthetic scores~\cite{aesthetic}.
For the prompt pairs generated by the mentioned Approaches i) and ii), we use the aesthetic score model~\cite{aesthetic} to score the images, as DiffusionDB already contains the images corresponding to the high-quality prompts.
For high-quality prompts generated by the mentioned Approach iii), we use the reward model $r_{aes}$ in Section \ref{reward-model} to compute the scores. 

We also consider prompts' consistency, calculate the text similarity~\cite{sbert} between low-quality and high-quality prompts in a pair, and filter out examples with low similarity. More details can be found in the Appendix \ref{data-details}.

\noindent\textbf{Statistics.}
We finally collect 143k prompt pairs as our training set. In addition, we randomly extract 2k entries from low-quality prompts as our testing set. For the training set, the average lengths of low-quality and high-quality prompts are 40.3 and 197.8, respectively, indicating that high-quality prompts contain more descriptions of details. More statistics can be found in Table \ref{statistics}.

\linespread{1.2}
\begin{table}[t]
\centering
\setlength{\tabcolsep}{3pt}{
\small
\begin{tabular}{lcccccc}

\toprule
\specialrule{0em}{1pt}{1pt}
 \textbf{Split} & \textbf{Source} & \textbf{Num} & \textbf{Aesthetic} & \textbf{PC} & \textbf{ALLP} & \textbf{ALOP} \\ 
 \specialrule{0em}{1pt}{1pt} \hline \specialrule{0em}{1pt}{1pt}

\multirow{4}{*}{Train} & All & 143k & 6.22 & 0.71 & 40.3 & 197.8 \\
& Summary & 134k & 6.23 & 0.71 & 39.8 & 194.5 \\
& Generation & 2k & 5.70 & 0.76 & 52.4 & 501.4 \\
& Caption & 7k & 6.23 & 0.67 & 44.9 & 177.7 \\

\specialrule{0em}{1pt}{1pt} \hline \specialrule{0em}{1pt}{1pt}
Test & - & 2k & - & - & 36.7 & - \\

\specialrule{0em}{1pt}{1pt}
\bottomrule
\end{tabular}
}
\normalsize
\linespread{1}
\caption{Dataset statistics. Note that, PC, ALLP and ALHP denote the prompt consistency (i.e., text similarity), the average lengths of low-quality and the high-quality prompts, respectively.}
\label{statistics}
\end{table}

\section{The~\emph{BeautifulPrompt} Model}

\begin{figure}[ht]
\centering
\includegraphics[width=.5\textwidth, keepaspectratio]
{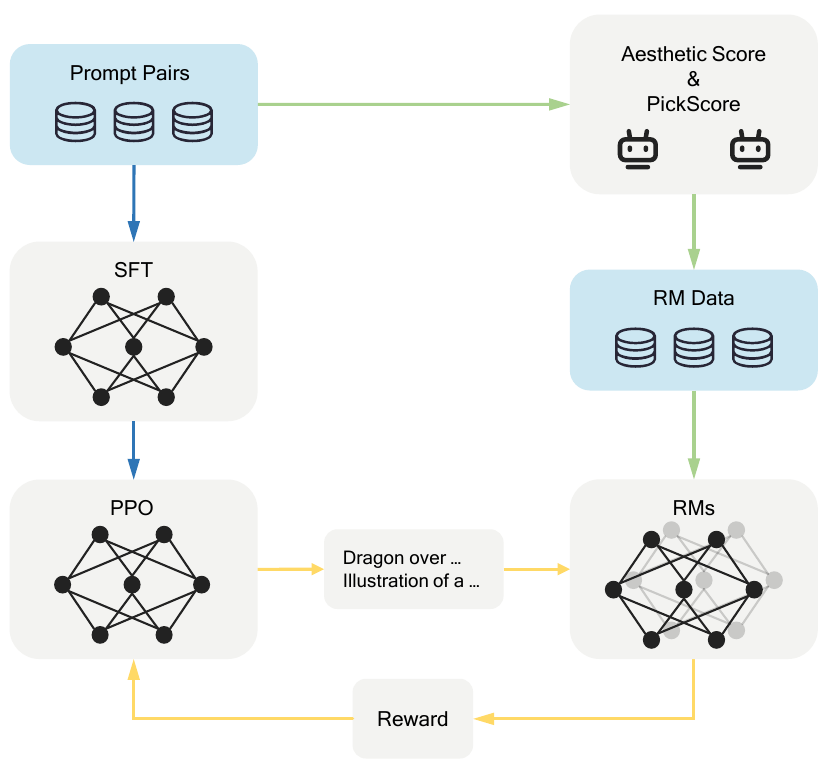}
\caption{The three steps of training~\emph{BeautifulPrompt}. The color of the arrows indicates three different stages.}
\label{method}
\end{figure}

Inspired by InstructGPT~\cite{instructgpt} and ChatGPT, in this section, we introduce the~\emph{BeautifulPrompt} training scheme in detail,
which contains three stages (Supervised Fine-tuning, Reward Modeling training and Reinforcement Learning), as shown in Figure \ref{method}.

\subsection{Supervised Fine-tuning (SFT)}

Given a dataset of prompt pairs $D=\{(\mathbf{x},\mathbf{y})\}$, containing pairs of low-quality prompts $\mathbf{x}$ and high-quality prompts $\mathbf{y}$, we fine-tune a decoder-only language model to output a high-quality prompt of tokens $\mathbf{y} = \{y_1, ..., y_n \}$ with a given instruction and a low-quality prompt $\mathbf{x}$.
We use the auto-regressive language modeling objective to maximize the following likelihood~\cite{gpt2}:
\begin{equation*}
\mathcal{L}_{sft} = - \sum_i \log P(y_i\mid \mathbf{x}, y_1, ..., y_{i-1}).
\end{equation*}

\subsection{Reward Modeling (RM)}
\label{reward-model}
Human feedback instructs the training of Large Language Models (LLMs) with promising results~\cite{instructgpt}. However, this requires extensive and tedious labor efforts. \citet{cai} propose to use AI models to instruct the training of LLMs.
Taking inspiration from this and considering that our final generated prompts $\mathbf{y}$ are used for drawing, we propose RLVAIF: a method that incorporates visual feedback into the training of language models, thereby avoiding the cost of expensive human labeling.

We focus on the quality of the final generated image and its similarity to the low-quality prompt $\mathbf{x}$. Therefore, we consider PickScore~\cite{pickscore} and the aesthetic score~\cite{aesthetic} as our visual AI feedback to 
train reward models to fit these scores.

Briefly, PickScore~\cite{pickscore} is a preference model trained on a large dataset of text-to-image prompts and real user preferences.
In order to reduce the impact of random seeds on the quality of the images generated by the TIS model, we use 8 different random seeds to generate images and average the results. The calculated averaged PickScore $\mathbb{PS}$ is used as the ground truth to train the reward model. The loss function is:
\begin{equation*}
\mathcal{L}_{ps} = - \frac{1}{N}\sum_i^N \text{MSE}(r_{ps}(\mathbf{x}, \mathbf{y}), \mathbb{PS}), 
\end{equation*}
where $r_{ps}(\mathbf{x}, \mathbf{y})$ is the scalar output of the reward model for the prompt pair $(\mathbf{x},\mathbf{y})$. MSE is the Mean Squared Error. $N$ is the total number of samples.

The aesthetic score model~\cite{aesthetic} is trained to predict the rating that people give when asked ``how much do you like this image on a scale from 1 to 10''. 
Similarly, a reward model is trained to fit the corresponding prompts from the images to the aesthetic scores $\mathbb{AES}$:
\begin{equation*}
\mathcal{L}_{aes} = - \frac{1}{N}\sum_i^N \text{MSE}(r_{aes}(\mathbf{y}), \mathbb{AES}),
\end{equation*}
where $r_{aes}(\mathbf{y})$ is the scalar output of the reward model.
Finally, we use $\alpha$ as a balancing factor to combine the scores of the two reward models as the final reward $r(\mathbf{x}, \mathbf{y})$:
\begin{equation*}
r(\mathbf{x}, \mathbf{y}) = \alpha \cdot r_{ps}(\mathbf{x}, \mathbf{y}) + (1 - \alpha) \cdot r_{aes}(\mathbf{y}).
\end{equation*}

\begin{table*}[ht]
\centering
\begin{small}
\begin{tabular}{lccccc}
\toprule
\textbf{Method} &\textbf{PickScore} & \textbf{Aesthetic Score} & \textbf{HPS} & \textbf{CLIP Score} & \textbf{Avg. Score} \\
\hline
Original & 20.74 & 5.50 & 0.197 & \textbf{0.27} & 0.57\\
\hline
MagicPrompt& 20.11 & 5.79 & 0.193 & 0.22 & 0.07\\
ChatGPT & 20.73 & 5.92 & 0.198 & 0.25 & 0.59\\
\hline
\emph{BeautifulPrompt} (SFT only)  & 20.42 & 6.03 & 0.197 & 0.23 & 0.39 \\
\emph{BeautifulPrompt} (Full implementation)  & \textbf{20.84} & \textbf{6.52} & \textbf{0.203} & 0.24 & \textbf{0.85} \\
\bottomrule
\end{tabular}
\end{small}
\caption{Results on the testing set. The average score is calculated with all scores normalized into [0,1].
``Original'' refers to the method that directly sends the original prompts to Stable Diffusion without modification.}
\label{result}
\end{table*}

\subsection{Reinforcement Learning}
As the collected dataset inevitably contains some noise, for example, the consistency between low-quality prompts and the corresponding high-quality prompts is relatively low, the performance of the supervising trained model $\rho$ can be unsatisfactory. To further improve the model performance, we initialize a policy $\pi = \rho$, and then fine-tune $\pi$ to perform the task using reinforcement learning. We leverage the Proximal Policy Optimization (PPO)~\cite{ppo} algorithm to directly optimize the expected reward:
\begin{equation*}
\mathbb{E}_{\pi} [r] = 
\mathbb{E}_{\mathbf{x}\sim D, \mathbf{y} \sim \pi(\cdot \mid \mathbf{x})} 
[r(\mathbf{x}, \mathbf{y}) - \beta\cdot\log \frac{\pi(\mathbf{y}\mid\mathbf{x})}{\rho (\mathbf{y}\mid\mathbf{x})}],
\end{equation*}
where $\beta$ is the Kullback-Leibler (KL) penalty coefficient. It prevents the policy from moving too far from $\rho$. Following \citet{rlhf}, we adopt an adaptive KL penalty here.

\section{Experiments}

\noindent\textbf{Training Settings.}
We use the pre-trained checkpoint of BLOOM~\cite{bloom} (1.1B parameters with 24 transformer layers) as the backbone.\footnote{
We choose a relatively small version of BLOOM as the backbone to ensure the high inference speed of online deployment to support real-world applications. In addition, we find that the 1.1B model is sufficiently large to accomplish our task effectively with good results. 
}
The BFLOAT16 formats are leveraged to save GPU memory and speed up training. 
For the SFT and RM stages of training, we set the batch size to 64, the maximum length to 384, and the learning rate to 1e-5 with warmup and cosine decay. We find that proper over-fitting benefits PPO training, so we set the SFT training epoch to 4 and the weight decay to 0.                                                                                                                                                                                                                                                                                                                                                                     
For PPO training, we set the learning rate to 5e-6, $\alpha$ to 0.7, the batch size to 32, the initial KL coefficient to 0.05, the training step to 5000, and freeze two-thirds of the parameters.
All the experiments are implemented in PyTorch and run on a single server with NVIDIA Tesla A100 GPUs.

\noindent\textbf{Baselines.}
We consider two strong baselines: MagicPrompt and ChatGPT. MagicPrompt is a popular automatic prompt completion model trained from 80,000 pieces of data crawls from Lexica.ai (refer to related work). ChatGPT is almost the most powerful general-purpose LLM and serves as a human-level prompt engineer here.

\noindent\textbf{Evaluation Protocols.}
Systematically evaluating the goodness of a prompt engineer is a challenging task. One of the most straightforward methods is to evaluate the images generated by the prompts that models produce. We use Stable Diffusion 1.5\footnote{\url{https://huggingface.co/runwayml/stable-diffusion-v1-5}} to generate images and calculate PickScore~\cite{pickscore}, the aesthetic score~\cite{aesthetic}, HPS~\cite{hps} and CLIP score~\cite{clip} for the images and the original prompts. 
In addition, we conduct a human evaluation experiment on 200 randomly selected examples from the testing set. Given the raw prompts, we ask 10 human experts to pick the most desirable images generated by the different methods and report the win rates of~\emph{BeautifulPrompt} compared against other methods.
\footnote{Refer to the user interface in Appendix~\ref{hpe}.}

\subsection{Overall Results}

\begin{figure}
\centering
\includegraphics[width=.48\textwidth, keepaspectratio]
{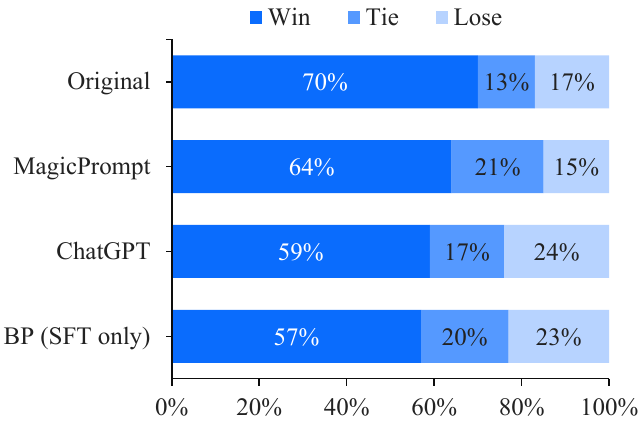}
\caption{Results of human preference evaluation (i.e., win/lose/tie rates of our method against others). ``BP'' is short for~\emph{BeautifulPrompt}. }
\label{human-result}
\end{figure}

From Table \ref{result}, our method consistently outperforms the other baselines in most scores.
As the CLIP score reflects the semantic consistency between the text and image, it is natural that sending the original prompts to Stable Diffusion unchanged obtains the highest score. Our method does not decrease the CLIP score to a large extent, showing that~\emph{BeautifulPrompt} well preserves the semantics of the original input prompts.
As shown in Figure \ref{human-result}, the human evaluation experiment shows the superiority of our approach, with a win rate of over 57\% against all other baselines. 

\subsection{Detailed Analysis}

\begin{figure}
\centering
\includegraphics[width=.48\textwidth, keepaspectratio]
{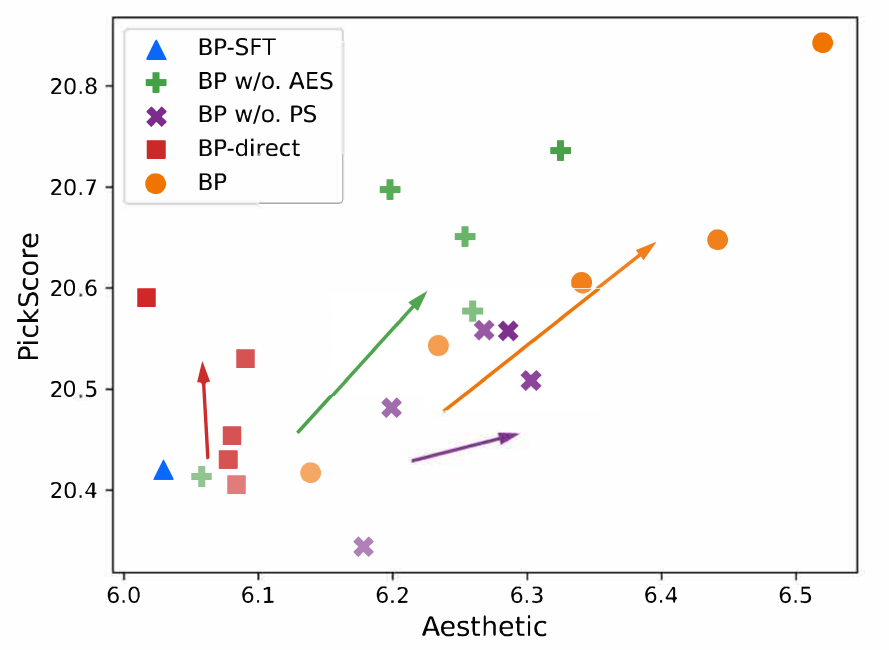}
\caption{Aesthetic-PickScore plot for~\emph{BeautifulPrompt} and its variants. ``BP'' is short for~\emph{BeautifulPrompt}. 
We visualize checkpoints every 1000 training steps. The color gradually darkens as the number of training steps increases and the
arrows indicate the training direction. For both scores, higher numbers are better.}
\label{as}
\end{figure}

\noindent\textbf{Ablation Study.}
Figure \ref{as} illustrates the training process using one reward model alone, two reward models, and directly using existing models to score the images as the reward.
Using $r_{ps}$ alone can drive an increase in aesthetic score, while using $r_{aes}$ alone does not drive an increase in PickScore. This is consistent with the finding that PickScore reflects real human preferences, incorporating various factors such as image aesthetics, text-image matching, etc~\cite{pickscore}. Combining the two rewards allows for more rapid and stable growth of both metrics and makes the training process more stable. 
The training process is unstable and the gains obtained are small when we directly use the models~\cite{aesthetic,pickscore} to compute rewards on the generated images instead of additionally training the reward models. Consistent with \citet{rlhf}, we observe that reward models need to understand languages to better guide training.

\begin{figure*}[ht]
\centering
\includegraphics[width=0.95\textwidth, keepaspectratio]
{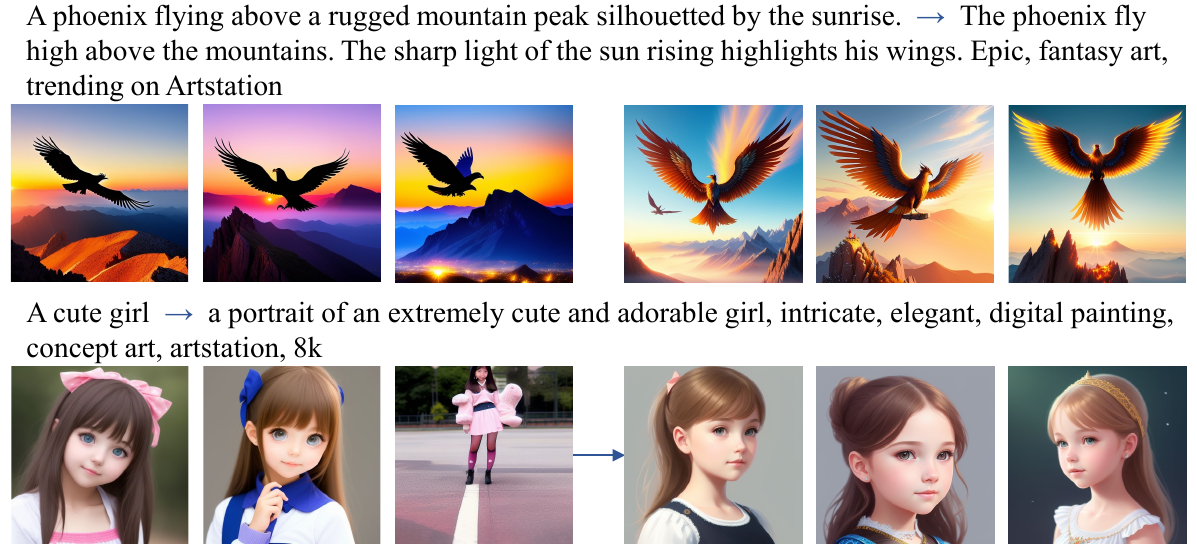}
\caption{
Comparing the qualities of images generated from the original prompts (left) with those from the prompts generated by~\emph{BeautifulPrompt} (right). The underlying TIS model is Delibrate.}
\label{imgs2}
\end{figure*}

\noindent\textbf{Is~\emph{BeautifulPrompt} Transferable?}
We further explore the transferability of~\emph{BeautifulPrompt} to the other diffusion-based TIS models.
Consider the popular model Deliberate\footnote{\url{https://huggingface.co/XpucT/Deliberate}}. As shown in Figure \ref{imgs2}, although Deliberate already performs well in most vanilla prompts,~\emph{BeautifulPrompt} is still able to make Deliberate generate more beautiful images in most cases.
This shows~\emph{BeautifulPrompt} can also be applied to other TIS models.
More examples can be found in the Appendix \ref{cases}.

\section{Industrial Application}

In this section, we briefly discuss how our model benefits users in real-world applications.
Currently, we have integrated~\emph{BeautifulPrompt} into a cloud-native AI platform (Platform of Artificial Intelligence, Alibaba Cloud\footnote{\url{https://www.alibabacloud.com/product/machine-learning}}) to assist users (especially designers and art workers) to create and edit artistic images based on a variety of Stable Diffusion-style models, together with other modules such as LoRA~\cite{lora} and ControlNet~\cite{controlnet}.
Users can freely perform any types of image generation and editing operations through WebUI. During any operation, users can invoke a~\emph{BeautifulPrompt} helper plug-in to assist the design or art creation process.
In addition, based on the Query Per Second (QPS) requirements and the system workload, our inference service can automatically scale to an adjustable number of machines on GPU clusters.

\begin{figure}
\centering
\includegraphics[width=.48\textwidth]
{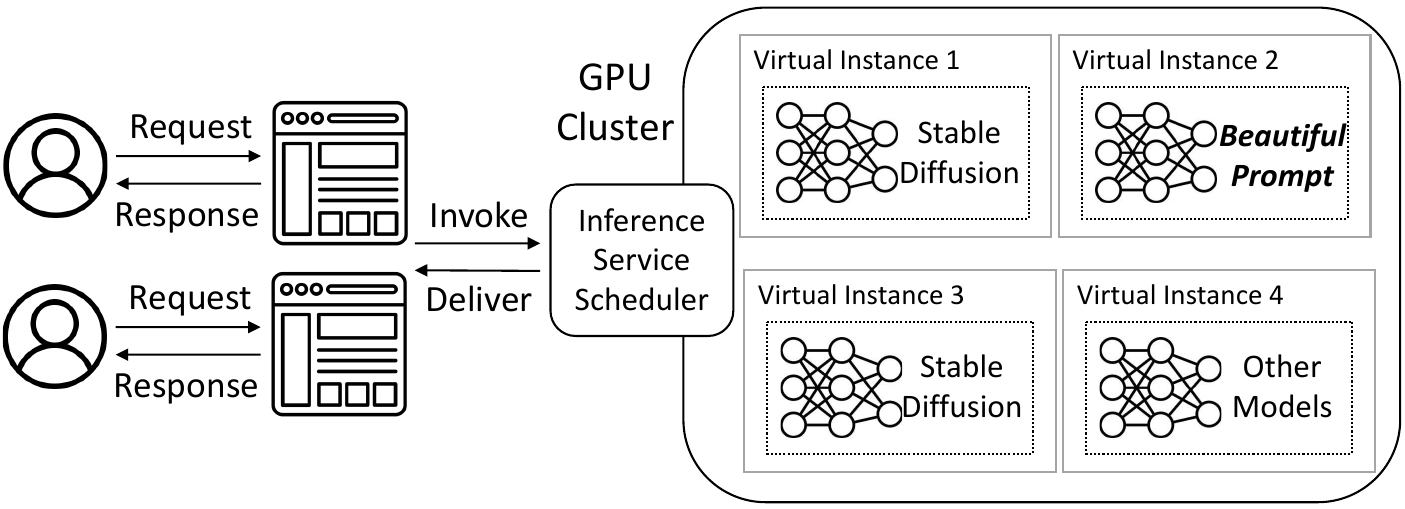}
\caption{Architecture of online deployment with~\emph{BeautifulPrompt} for text-to-image generation service. }
\label{eas}
\end{figure}

\section{Conclusion}
We propose a deep generative model named~\emph{BeautifulPrompt} to create high-quality prompts, which can be feed to Stable Diffusion-style models to produce more beautiful images. Specifically, we collect and release a new dataset for training prompt engineering models. A Reinforcement Learning with Visual AI Feedback technique is introduced to fine-turn the LLMs based on our dataset.
Extensive experimental results show that~\emph{BeautifulPrompt} outperforms existing methods in terms of both automatic and human evaluation.

\section*{Limitations}

Although~\emph{BeautifulPrompt} can generate more aesthetically pleasing images, limited by the training data, it sometimes ignores part of the information in the original prompts or generates meaningless prompts. In a few cases, the generated images can be semantically inconsistent with the original prompts, due to the auto-regressive and generative nature of language models.
These improvements are left to our subsequent work. In addition, multiple open-source models are used in our training data construction, and model training process, which may cause some degree of bias as well as error accumulation.

\section*{Ethical Considerations}
The techniques for training the~\emph{BeautifulPrompt} model presented in this work are fully methodological. Hence, there are no direct negative social impacts of our method. As for the model, to ensure that the generated contents are suitable for public release, we have also filtered out NSFW prompts from our training data.
However, since the generative process is difficult to control, it is possible (although not likely) for our model to create toxic contents.
We suggest that in our case,~\emph{BeautifulPrompt} should not be used to generate offensive or inappropriate images for people intentionally. Users should carefully deal with the
potential risks for online deployment.

\section*{Acknowledgements}

This work is partially supported by Alibaba Cloud Group through Research Talent Program with South China University of Technology.

\appendix

\begin{figure}[h]
\centering
\includegraphics[width=.49\textwidth, keepaspectratio]
{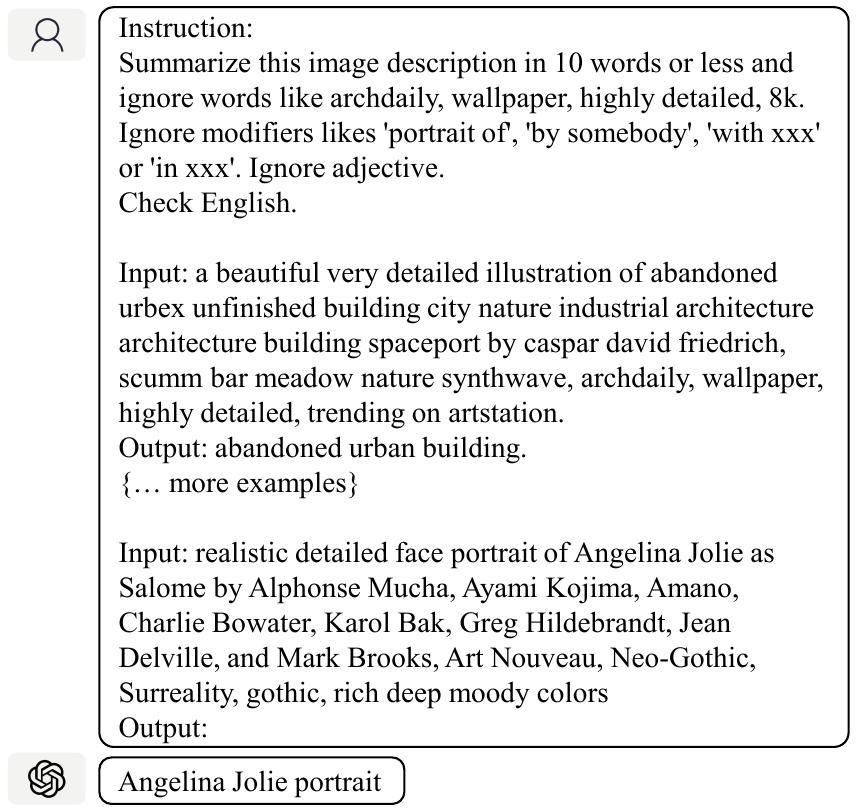}
\caption{An example of ``ChatGPT summary'' for data collection.}
\label{summary}
\end{figure}

\section{ChatGPT Templates}
\label{chatgpt}

\begin{figure}[h]
\centering
\includegraphics[width=.49\textwidth, keepaspectratio]
{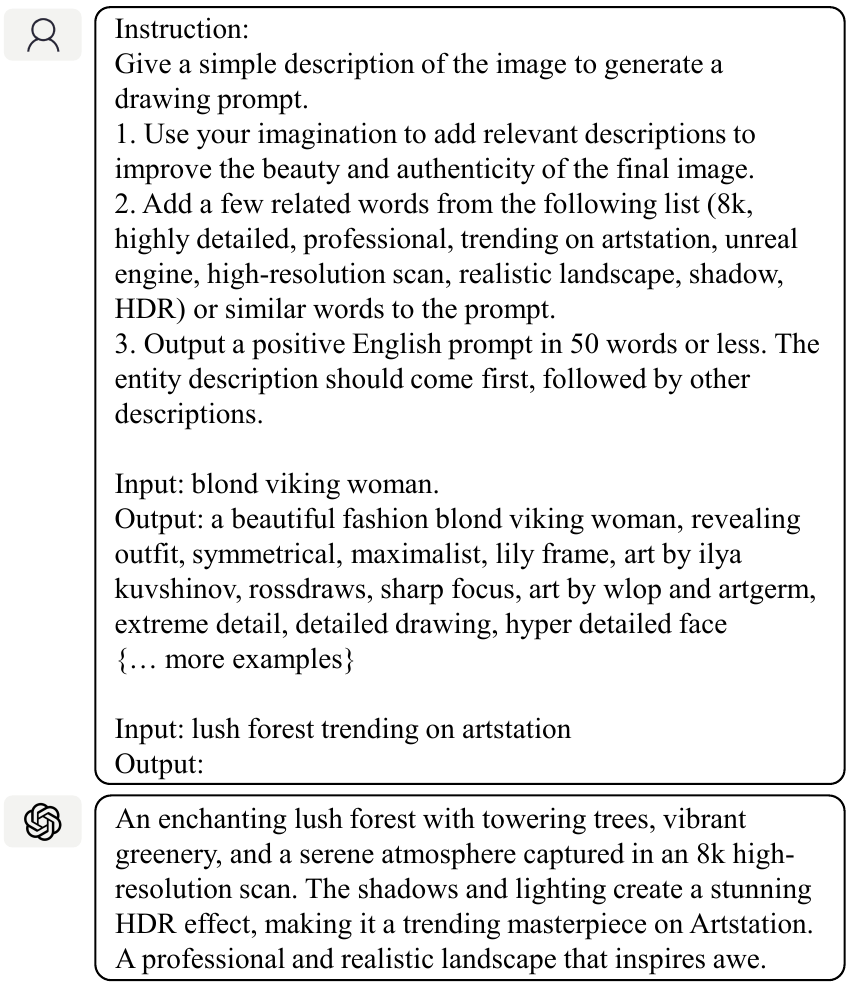}
\caption{An example of ``ChatGPT prompt generation'' for data collection.}
\label{generate}
\end{figure}

Figure \ref{summary} and Figure \ref{generate} show examples of using ChatGPT to generate part of the training set.

\begin{figure}[h]
\centering
\includegraphics[width=.5\textwidth, keepaspectratio]
{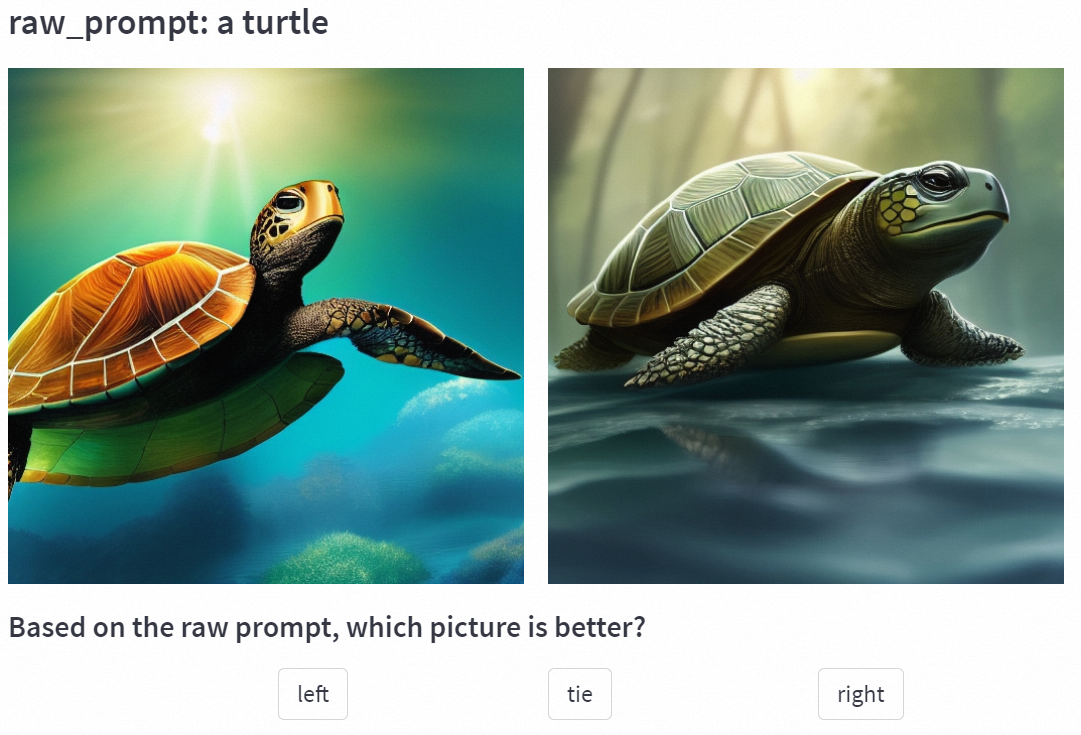}
\caption{Screenshot of the user inferface for the human evaluation experiment.}
\label{hpe-figure}
\end{figure}

\section{Data Post-processing Details}
\label{data-details}

For NSFW filtering, we use a trained NSFW classifier\footnote{\url{https://huggingface.co/michellejieli/NSFW_text_classifier}}. For consistency filtering, we first use the trained sentence encoder\footnote{\url{https://huggingface.co/sentence-transformers/all-mpnet-base-v2}} to obtain sentence representations and then compute their cosine similarity:
$$
\text{cos\_sim}(\mathbf{r}_x, \mathbf{r}_y)=\frac{\mathbf{r}_x^\top \mathbf{r}_y}{\parallel \mathbf{r}_x \parallel \cdot \parallel \mathbf{r}_y \parallel},
$$
where $\mathbf{r}_x$ and $\mathbf{r}_y$ are sentence representations of low- and high-quality prompts.

\section{Human Preference Evaluation}
\label{hpe}
Figure~\ref{hpe-figure} shows a screenshot of the human evaluation experiment.


\section{More Cases}
\label{cases}
\begin{figure*}[ht]
\centering
\includegraphics[width=\textwidth, keepaspectratio]
{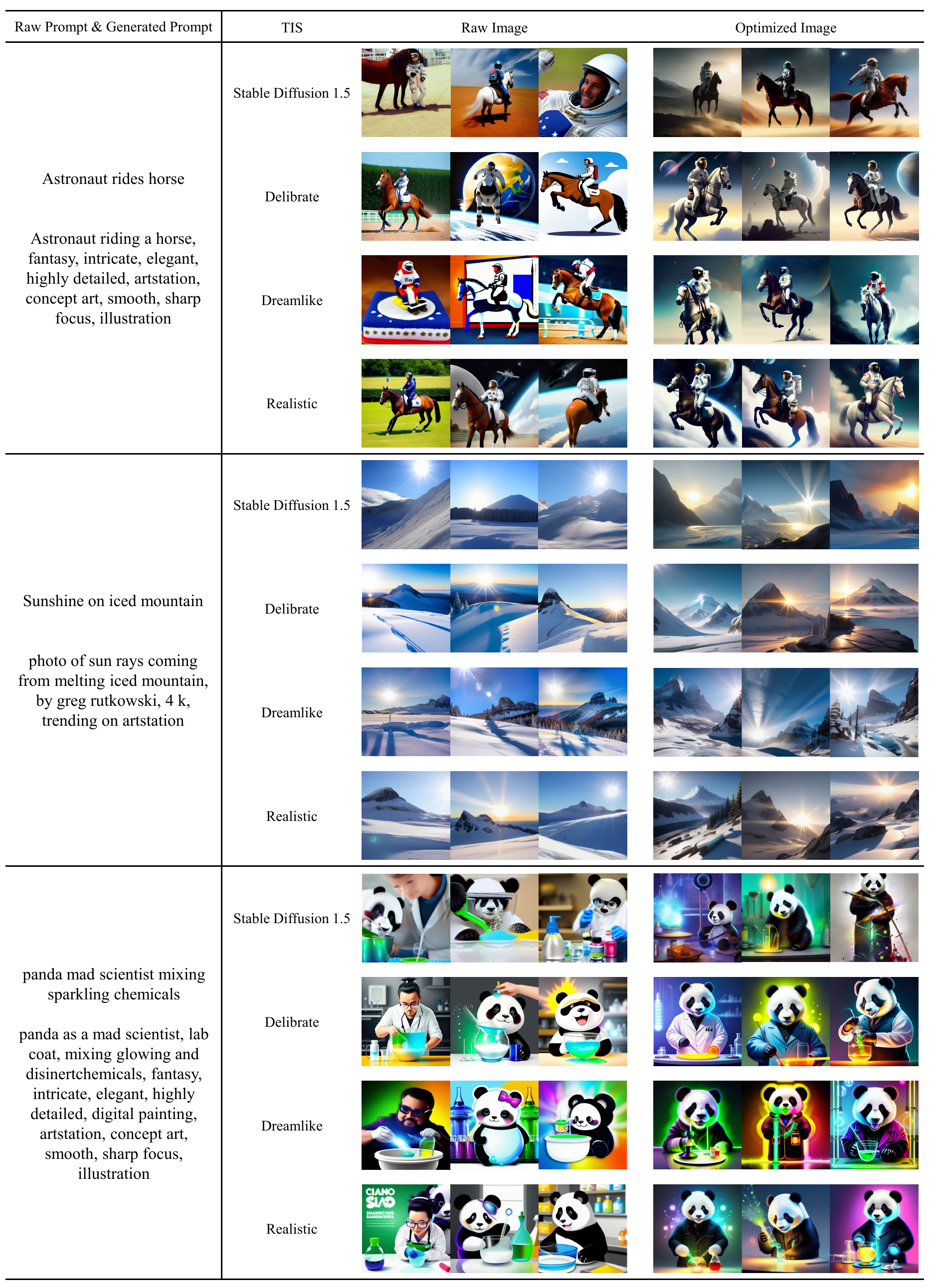}
\caption{Examples of images generated by various Stable Diffusion-style models w/ and w/o~\emph{BeautifulPrompt}.}
\label{imgs4}
\end{figure*}

In Figure \ref{imgs4}, we apply \emph{BeautifulPrompt} to more Stable Diffusion-style models (i.e., Stable Diffusion 1.5, Delibrate, Dreamlike\footnote{\url{https://huggingface.co/dreamlike-art/dreamlike-photoreal-2.0}} and Realistic\footnote{\url{https://huggingface.co/SG161222/Realistic\_Vision\_V1.4}}).

\end{document}